\documentclass[10pt,twocolumn,journal]{IEEEtran}
\usepackage[T1]{fontenc}
\usepackage[latin9]{inputenc}
\usepackage{graphicx}
\usepackage{balance}
\usepackage{hyperref}
\makeatletter
\usepackage{xcolor}

\makeatother

\begin{document}
\title{Opportunities of Federated Learning in Connected, Cooperative and Automated Industrial Systems}
\author{Stefano Savazzi, Monica Nicoli, Mehdi Bennis, Sanaz Kianoush, Luca
Barbieri\thanks{\textcopyright  2021 IEEE.  Personal use of this material is permitted.  Permission from IEEE must be obtained for all other uses, in any current or future media, including reprinting/republishing this material for advertising or promotional purposes, creating new collective works, for resale or redistribution to servers or lists, or reuse of any copyrighted component of this work in other works.} \thanks{S. Savazzi and S. Kianoush are with the Institute of Electronics, Computer and Telecommunication Engineering (IEIIT) of Consiglio Nazionale delle Ricerche (CNR), Milano, Italy.} \thanks{M. Nicoli and L. Barbieri are with Politecnico di Milano DIG and DEIB department, Milano, Italy.} \thanks{M. Bennis is with the Centre for Wireless Communications, University of Oulu, Finland.} \thanks{The paper has been accepted for publication in the IEEE Communications Magazine. The work is supported by the CHIST-ERA European projects RadioSense (Wireless Big-Data Augmented Smart Industry) and CONNECT (Communication-aware Dynamic Edge Computing). It is also supported by the Project BASE5G (Broadband InterfAces and services for Smart Environments enabled by 5G technologies), funded by the Italian Lombardy Regional Government under the grant POR-FESR 2014-2020, ID 1155850. The current arXiv contains an additional Appendix that describes the MIMO radar dataset for the setup of Fig. 5. Sample dataset published on IEEE Data Port http://dx.doi.org/10.21227/0wmc-hq36.}}
\maketitle
\begin{abstract}
Next-generation autonomous and networked industrial systems (i.e.,
robots, vehicles, drones) have driven advances in ultra-reliable,
low latency communications (URLLC) and computing. These networked
multi-agent systems require fast, communication-efficient and distributed
machine learning (ML) to provide mission critical control functionalities.
Distributed ML techniques, including federated learning (FL), represent
a mushrooming multidisciplinary research area weaving in sensing,
communication and learning. FL enables continual model training in
distributed wireless systems: rather than fusing raw data samples
at a centralized server, FL leverages a cooperative fusion approach
where networked agents, connected via URLLC, act as distributed learners
that periodically exchange their locally trained model parameters.
This article explores emerging opportunities of FL for the next-generation
networked industrial systems. Open problems are discussed, focusing
on cooperative driving in connected automated vehicles and collaborative
robotics in smart manufacturing. 
\end{abstract}

\IEEEpeerreviewmaketitle{}

\section{Introduction}

The rapid transformation of industrial systems, driven by the digitization
and 5G communication evolution, has led to extensive research initiatives
on manufacturing and automotive verticals. These include, for example, Industry 4.0
(I4.0), at European level, the European Factories of the Future Research
Association (EFFRA, effra.eu), the 5G Alliance for Connected Industries
and Automation (5g-acia.org) and the 5G Automotive Association (5gaa.org).
The envisioned smart industrial systems rely on networked \emph{machines} with increasing level of intelligence and autonomy,
moving far beyond traditional low-cost, low-power sensors. According
to industrial requirements, such machines are required to support:

1) autonomous and adaptive decision making in dynamic situations with mobile operators/equipment, device-less human-machine interfaces and time-varying environments;

2) big-data-driven training of large-size machine learning (ML) models
for decision-making;

3) ultra-reliable and low-latency communications (URLLC) \cite{bennis}
for mission-critical device-to-device (D2D) operations.

Networked and cooperative intelligent machines have recently opened new research opportunities that target the integration of distributed ML tools with sensing, communication and decision operations. Cross-fertilization
of these components is crucial to enable challenging collaborative
tasks in terms of safety, reliability, scalability and latency.

Among distributed ML techniques, federated learning (FL) \cite{kone1,survey}
has been emerging for model training in decentralized wireless systems. Model
parameters, namely weights and biases in deep neural network (DNN)
layers, are optimized collectively by cooperation of interconnected
devices, acting as distributed learners. In contrast to conventional
edge-cloud ML, FL does not require to send local training data to the server, which may be infeasible in mission critical settings with extremely low latency and data
privacy constraints.

The most popular FL implementation, namely federated averaging \cite{kone1},
alternates between the computation of a \emph{local model} at each
device and a round of communication with the server for learning of
a \emph{global model}. Local models are typically obtained by minimizing
a local loss function via Stochastic Gradient Descent (SGD) steps
\cite{survey}, using local training examples and target values.

\begin{figure}[!t]
\center\includegraphics[scale=0.57]{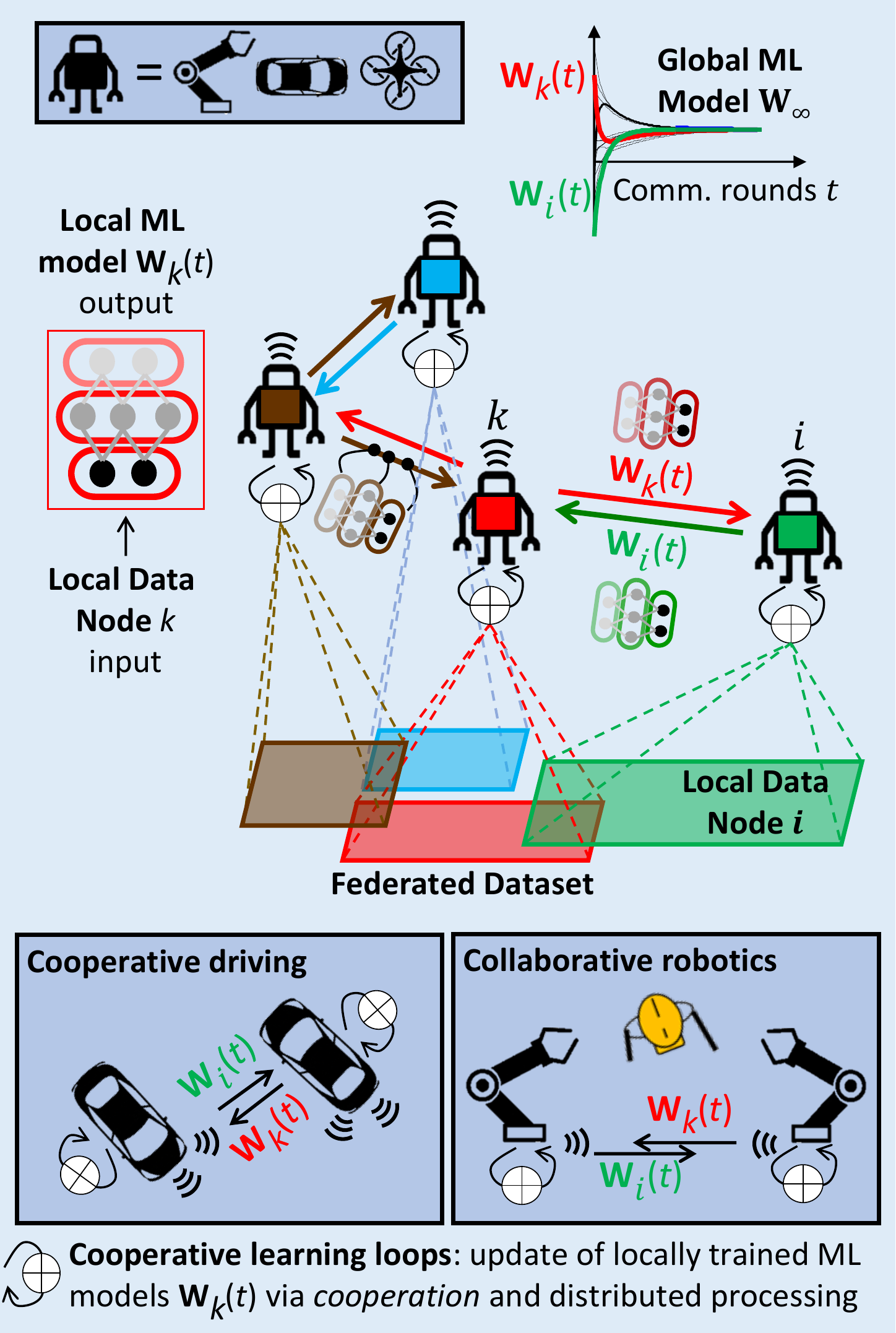} %\par\end{centering}
 \protect\caption{\label{fed} Decentralized learning with federated datasets. Examples in mission critical control applications: collaborative industrial robotics and cooperative automated vehicles.}
\end{figure}

Federated averaging is privacy-preserving by design, as it keeps the training
data on-device. However, it still leverages the server-client architecture,
which might not be robust to data poisoning attacks and scalability needs.
Overcoming this issue mandates moving towards fully decentralized FL
solutions relying \emph{solely} on local processing and cooperation among end machines. As shown in Fig. \ref{fed}, the device sends its local ML model parameters to neighbors and receives in
return the corresponding updates. Next, it improves
its local parameters by fusing the received contributions. This procedure
continues until convergence.

The article addresses the opportunities of emerging distributed FL
tools specifically tailored for systems characterized by autonomous
industrial components (vehicles, robots). FL is first proposed as
an integral part of the sensing-decision-action loop. Next, novel decentralized FL tools and emerging research
challenges are highlighted. The potential of FL is further elaborated with
considerations primarily given to mission critical control operations in the field of cooperative automated vehicles and densely interconnected robots. Analysis with real data on a practical usage
scenario reveals FL as a promising tool underpinned by URLLC communications.

\section{Sensing, decentralized learning and communication co-design}

\begin{figure}[!t]
\center\includegraphics[scale=0.48]{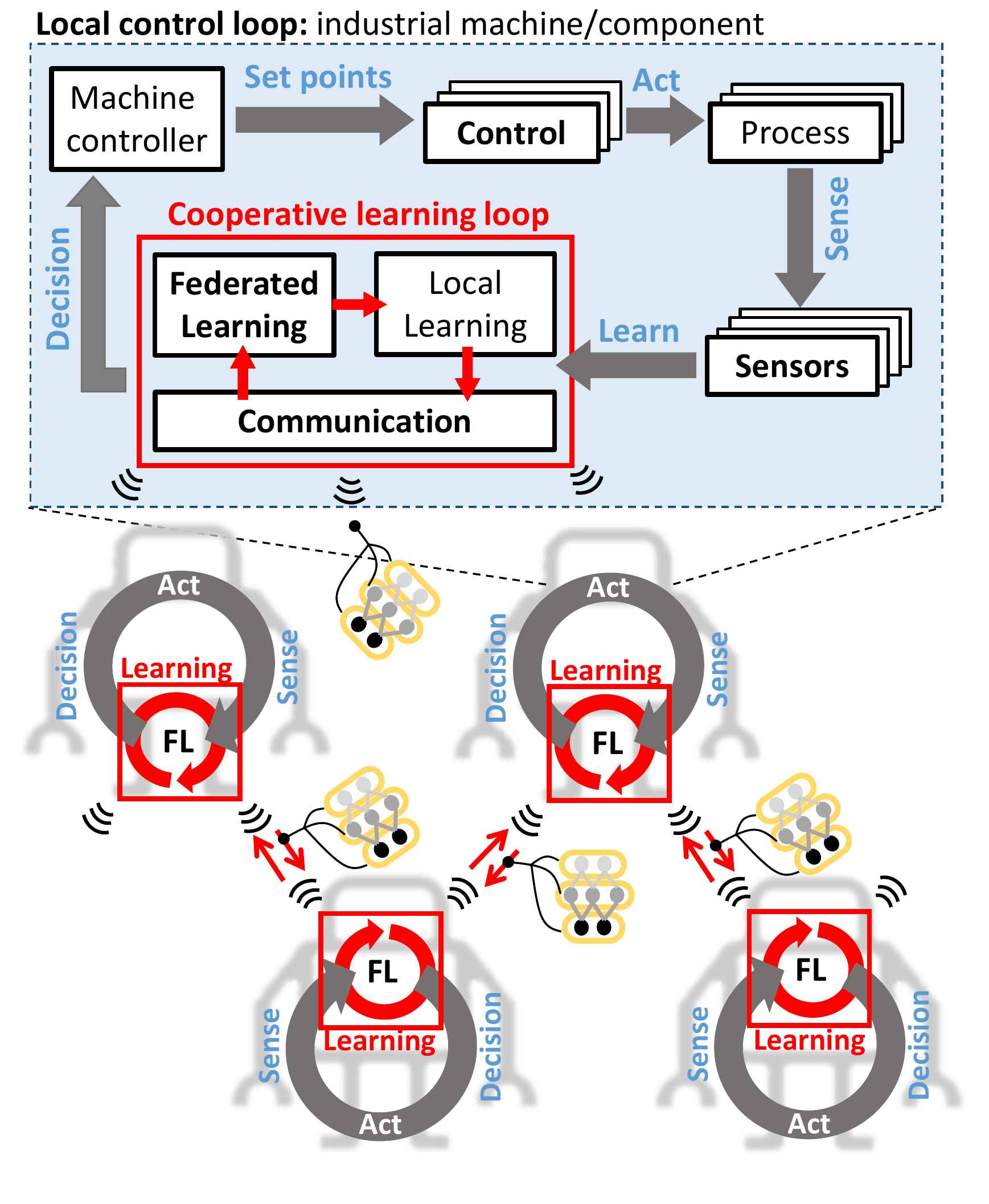} %\par\end{centering}
 \protect\caption{\label{fed-1} Decentralized FL in the sensing-decision-action loop.}
\end{figure}

Decentralized FL solutions have great potential for industrial 5G
and beyond 5G (B5G) verticals. In automated industrial processes, decentralized FL imbues intelligence directly into the end machines, which become smart cooperative agents. Fig.
\ref{fed-1} depicts a schematic of a cooperative and automated multi-agent
industrial system. It consists of connected machines performing collaborative
tasks, and integrating ML model training within the sensing-decision-action loop. The ML model outputs might be scenario-dependent predictions
of a physical process, or rather value functions to be used for policy
improvements (i.e., reinforcement learning). Outputs are fed to the
machine controller for local decisions or actuations. Training of
the ML model calls for highly efficient knowledge discovery operations based on the overall training data collected
by \emph{all} the machines performing the same task (federated dataset),
rather than local data only. Recent advancements of FL constitute
a first but significant step towards \emph{collaborative learning}
and, particularly, the understanding of how ML could be distributed
over networked devices, without centralized orchestration. Collaborative
learning underpins local decisions and allows the networked machines
to augment their ML model by sharing ego knowledge. As part of the
sensing-decision-action loop, emerging FL tools are expected to target
three fundamental challenges:

1) get over the restrictions of server-client architectures and movements of large, unstructured, raw datasets over D2D wireless
links, in favor of (typically) sparse ML model parameters exchange;

2) optimally balance opportunistic (ego) learning based on local training data,
and collaborative learning (leveraging neighbor's experience), with
the goal of steering convergence;

3) learn and re-train continuously over URLLC links to adapt to changes
in the data distribution, environment, process or situation.

\section{Decentralized FL: emerging trends}

Decentralized FL alternates the mutual exchange of local ML models
$\mathbf{W}_{k}(t)$ with the on-device minimization of a local loss
function. The goal is to promote convergence to a global model $\mathbf{W}_{\infty}$
that minimizes a global loss, decomposed into the sum of local losses.
Fully distributed FL implementations run on top of D2D networks characterized
by arbitrary connectivity graphs. The collaborative learning is based on Decentralized Stochastic Gradient Descent (DSGD), which guarantees convergence under strong convexity assumptions of local loss functions \cite{incentive} and networks with doubly stochastic adjacency matrix \cite{survey}. DSGD alternates on-device SGD steps to obtain $\mathbf{W}_{k}(t)$,
with the mutual exchange of model parameters to steer convergence.
Mutual exchange is regulated by gossip \cite{GOSSIPGRAD}, consensus
or diffusion algorithms \cite{cfa}-\cite{diffusion}.

\subsection{Consensus, diffusion and gradient negotiations}

In consensus based approaches, \cite{GOSSIPGRAD}-\cite{cfa}, federated
nodes exchange local ML model parameters and update them sequentially
by distributed averaging. For the mutual exchange of models, nodes
might select random, time-varying or optimized \cite{GADMM}-\cite{poor}
partners. In gossipgrad \cite{GOSSIPGRAD}, the selection policy creates
a virtual network where each cooperating agent is connected to, typically,
two other nodes. Convergence time and loss are ruled by the specific
choice of model update operations. Some updating strategies favor
fast convergence \cite{cfa} while others target model accuracy \cite{poor}.
Considering DNN model training, the neural network layers are also
typically learned separately and independently. Therefore, on-device
optimization and networking phases can run in parallel. Communication
overhead scales linearly with the model size and the number of cooperating
agents. However, non-independent-identically-distributed (non-IID)
federated datasets and large model size penalize the rate of convergence
\cite{survey}.

\begin{figure}[t]
\center\includegraphics[scale=0.38]{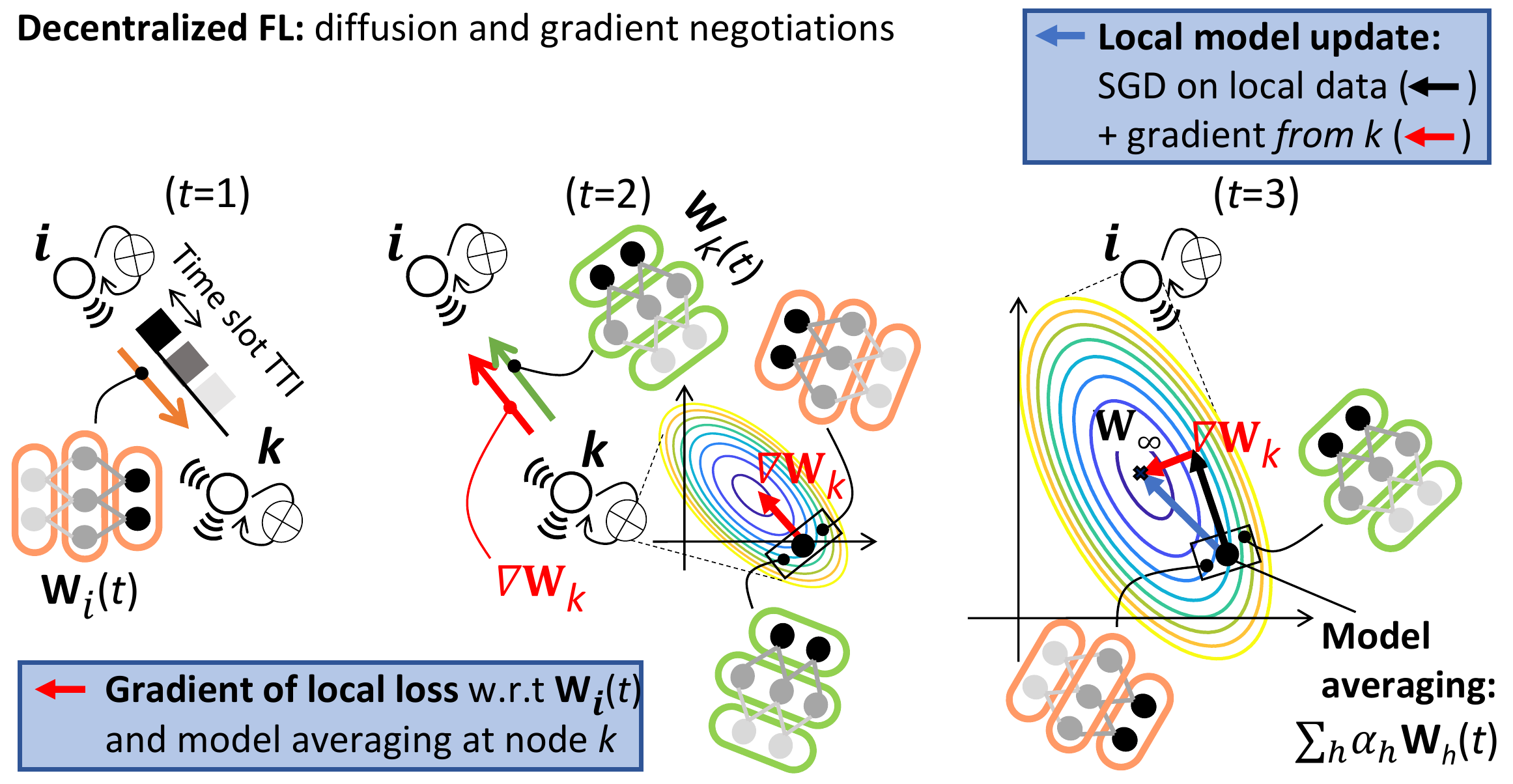} %\par\end{centering}
 \protect\caption{\label{exchange} Decentralized FL with diffusion: communication,
gradient negotiations ($t=1,2$) and computing rounds ($t=3$).}
\vspace{-0.4cm}
\end{figure}

Diffusion strategies incorporate a gradient negotiation phase,
described in Fig. \ref{exchange}, where devices exchange the information
about how neighbor models should be adjusted considering local data.
Diffusion virtually expands the local training data-set, and boosts
the convergence compared with gossip. Gradient exchange strategies
are often regulated by request-reply negotiation stages between neighbors
\cite{diffusion}. In the example of Fig. \ref{exchange}, the local
model $i$, $\mathbf{W}_{i}(t=1)$, is sent to the neighbor $k$ at
time $t=1$ to start a negotiation. The received model is used by
the neighbor $k$ to compute a gradient vector $\nabla\mathbf{W}$
using the local loss. Both the gradient vector and the local model
are then fed-back to the device $i$, that started the negotiation
($t=2$). On-device optimization ($t=3$) finally adjusts the local
model by combining the gradients obtained from local loss with those
received from the neighbor (combine-and-adapt \cite{diffusion}).
In \cite{cfa} the negotiation resorts to a two-stage asynchronous
procedure: convergence speed improves at the cost of larger communication
overhead. Exploring this trade off is currently an open challenge.

In addition to the above tools, distributed ledger technologies can also
be applied to decentralized training \cite{key-8} by 
validating clients model updates via a series of validation steps
(Proof-of-Work, or others). As a result, decentralized FL is transformed
into a market of expert model training nodes and validators. The development
of robust FL designs against data poisoning and adversarial manipulations
is still an open problem.

\subsection{Improving communication efficiency}

Reliable and low-latency D2D communications serve as the backbone
for distributed FL computations. Transmission of model parameters
must be extremely reliable with packet error rate down to $ 10^{-8}$
to prevent frequent retransmissions penalizing the convergence rate
\cite{cfa}. Transmission time intervals (TTI) should be aligned with
the data dynamics and the computation times required to perform SGD
and model adaptation, targeting $5$ ms and below \cite{bennis}.
The model size also affects communication design choices: DNN models
often used in industrial applications contain $>15$k parameters per
layer, usually extremely sparse. 

Although vanilla FL assumed noiseless
or rate-limited communications \cite{survey}, recently proposed \emph{digital}
and \emph{analog} designs quantify the effects of intermittent communications,
time-varying fading and interference. Digital implementations of FL require each device to communicate with peers over half-duplex links via time scheduled wireless
access. Popular solutions to limit the model size, and thus the communication
overhead, are quantization, sparsification and distillation \cite{GADMM}.
Sparsifying operators select a subset of informative ML parameters,
often improving also the model generalization.

More recently, analog FL and hybrid analog-digital implementations have been proposed for fast
and synchronous model averaging \cite{incentive}. Analog FL gets
over the restrictions of time scheduled access as it exploits the
superposition property of dense wireless transmissions when averaging
the neighbor models. Each device receives the superposition of the
ML models simultaneously transmitted by the neighborhood. When ML
parameters are sparse and sent as uncoded/uncompressed, Lasso type
recovery can be adopted for decoding and reconstruction \cite{digvanalog}.

Besides low-level communications, optimization of resource allocation has been considered to find the best trade-off between on-device computation (SGD
updates) and wireless channel use for parameters exchange \cite{feder}.
Optimized medium access control can also substantially reduce the
FL loss \cite{poor}: for example, scheduling should reward devices
having high quality data compared to those possessing few, or redundant
samples \cite{key-8}.

\subsection{Emerging research challenges}

The roll-out of new decentralized communication paradigms in B5G is
expected to bridge the gap between deep learning and wireless networking
research, raising at the same time unprecedented challenges. For example,
current FL designs generally ignore the underlying dynamics of the
network graph, the presence of intermittent communication links or
weakly connected components \cite{survey}. Research in this direction
should focus on balancing local data collection, model adaptations
and cooperation from selected agents. Digital, analog or hybrid implementations
should be considered based on devices' mobility, model size, computational
and bandwidth resources, as well as connectivity quality. Learning
an optimal policy for graph (and/or neighbor) selection while training
the ML model, namely learning \emph{simultaneously} the network graph
and the model, is also a promising direction.

Automated systems are often characterized by heterogeneous devices
performing distinct but related tasks. Collaborative learning of multiple
functions, namely multitask learning \cite{multitask}, will have
a relevant impact on several industrial applications. Understanding
when cooperation, in the form of consensus, or any decentralized agreement
protocol, can better steer convergence compared with opportunistic or ego
behaviors is an open problem of wide interest. The analysis must also
take into consideration the cost of distributed computations, the federated data distribution and the network designs for URLLC.

Using the I4.0 vision and emerging B5G verticals as guidelines, in what follows opportunities of FL are highlighted for the needs of selected mission critical control applications, namely cooperative automated driving and collaborative robotics.

\begin{figure}[!t]
\center\includegraphics[scale=0.54]{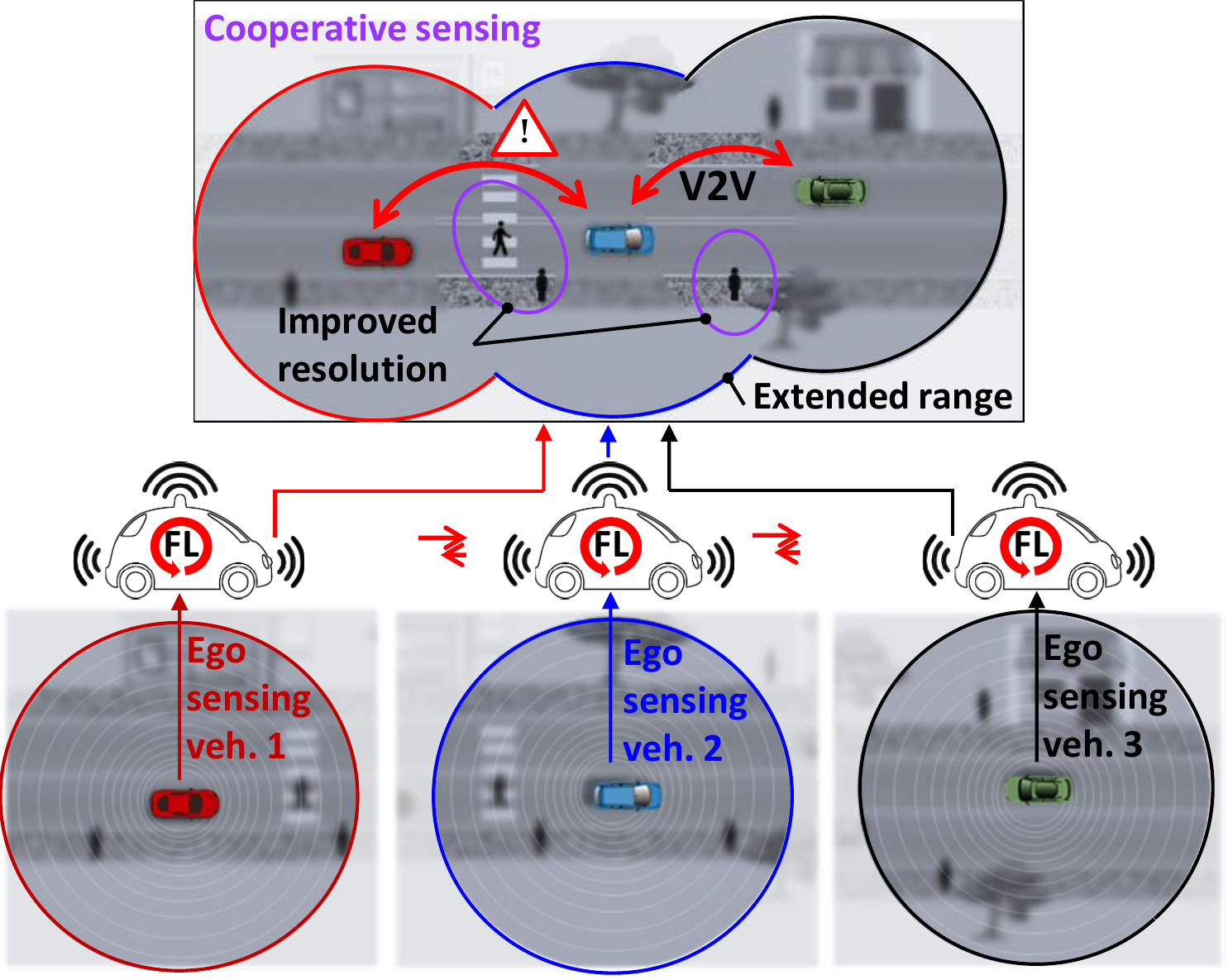} %\par\end{centering}
 \protect\caption{\label{vehicles}Distributed FL for cooperative perception by real-time
fusion of imaging data at different vehicles.}
\end{figure}

\section{Distributed intelligence for cooperative automated driving}

Vehicular URLLC leverage B5G connectivity to enable flexible vehicle-to-everything
(V2X) interactions with road infrastructure (V2I) and other vehicles
(V2V). Distributed ML over V2X networks will play a central role in
cooperative intelligent transportation system (C-ITS), enabling level
4-5 cooperative automated driving functionalities \cite{jointopt}.
Cooperative automated vehicles share maps of the driving environment
(Fig. \ref{vehicles}) using V2X URLLC to extend the range and resolution
of their ego imaging sensors (radar, camera, lidar). Beside improving detection
and localization of safety-related events, vehicles can also negotiate
the maneuvering and synchronize to a common mobility pattern, forming
tight autonomous-driving convoys and increasing traffic efficiency.
All the above scenarios are characterized by time-critical functions
that must be implemented on a closer-to-the-ground cloud, with part of the cooperative computational tasks performed locally, by pushing intelligence into the
vehicles rather than on the mobile edge cloud (MEC) \cite{jointopt}.

Decentralized FL techniques \cite{cfa} are promising solutions for
these time-sensitive applications. They require the vehicles to transmit smaller models which can be
aggregated at the road side unit (RSU) or by vehicles via ultra-low latency
($3$ ms) and highly reliable ($10^{-5}$) V2X connectivity. The exchange of local ML model parameters, rather than raw data, is expected to decrease the learning time, allowing to quickly react to unexpected events and take safety-critical decisions.
Domain-specific FL designs must also account for intermittent communications, time-varying network graphs and non-IID training datasets changing quickly over time, thus evolving according to vehicle motions, with speeds up to 250 km/h (3GPP TR 22.886), and the surrounding environment. In such cases, a transitory phase is expected where vehicles with outdated, or partially trained models, will coexist with highly-automated ones and benefit from their cooperation. The federation with fully equipped vehicles will assist lower-level vehicles to get an augmented vision of the driving environment, even if equipped with less accurate sensors. Balancing between centralized and decentralized FL implementations for energy optimization is also a main issue. Besides
cooperative driving, to improve safety and reliability, the vehicles
need also to learn the network latency in a distributed manner \cite{bennis},
whereby decentralized FL (and its variants) are instrumental. Open
problems further relate to the tension between local and global models, and
the impact of large numbers of vehicles.

Fig. \ref{vehicles} highlights a distributed ML-assisted cooperative
perception task where vehicles fuse their sensor data in a decentralized
implementation to get an extended vision of the environment. Fusing dynamics (i.e., global navigation satellite systems, GNSS, inertial measurement units, IMU) and imaging sensor data (i.e., lidar, camera, radar) from different vehicles improves
the location sensing accuracy making real-time responses feasible. 
A main challenge is the association of unlabeled imaging measurements
at different vehicles to jointly sensed features for cooperative simultaneous localization and mapping (SLAM).
Large data
volumes and computational complexity are also critical challenges.
Decentralized FL is a promising candidate, as it is able to learn
a common model for data association and fusion from local raw data,
limiting the V2X exchange to model parameters.

\begin{figure*}[!t]
\center\includegraphics[scale=0.47]{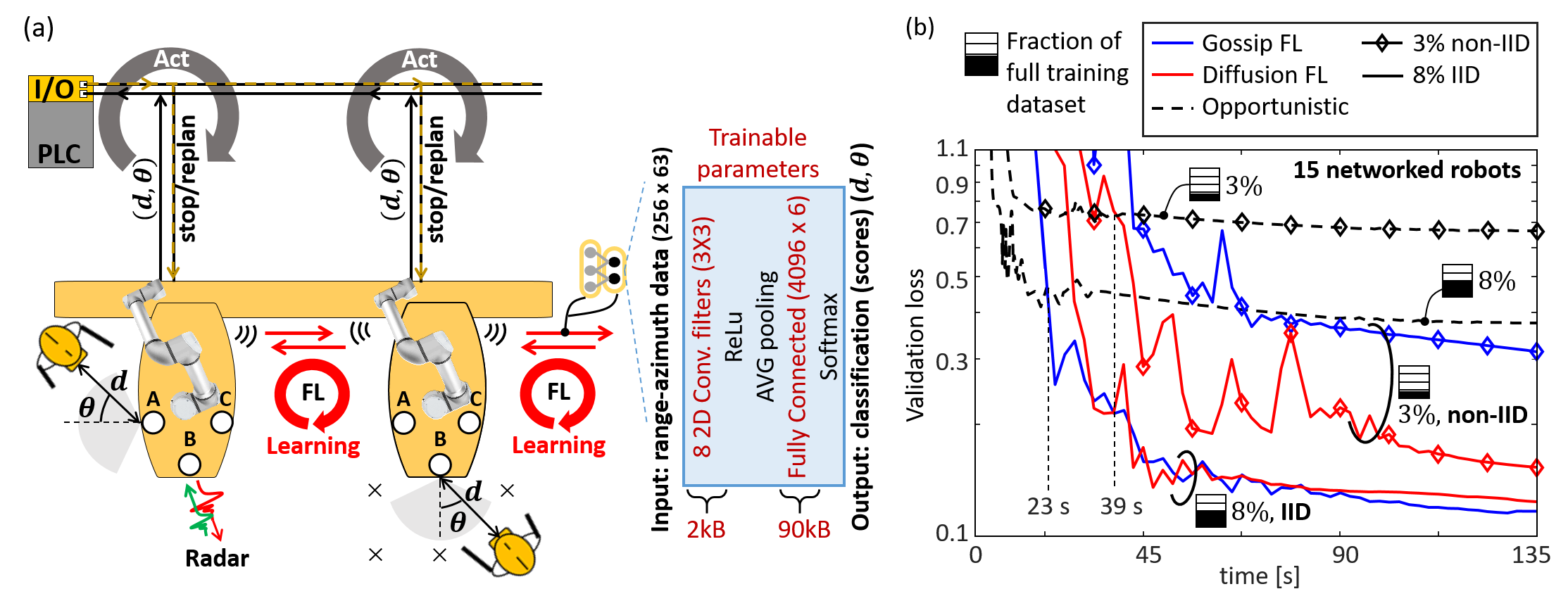} %\par\end{centering}
 \protect\caption{\label{fed-2} (a) Decentralized FL setup in a cobot environment: frequency-modulated continuous wave (FMCW), multiple-input-multiple output (MIMO) radar locations (A,B,C) and ML model based classification of subject positions; (b) FL loss for gossip (blue) and diffusion (red) vs. time and over a D2D network of $15$ robots. IID (no markers) and non-IID (diamond markers) federated datasets using the $8\%$ and $3\%$ of the full training dataset, respectively. D2D network with TTI=$3$ms, BLER=$10^{-9}$. Dashed lines refer to the opportunistic learning case.}
\end{figure*}

\section{Decentralized FL for cobots}

The development of collaborative robotics in advanced manufacturing environments (cobots) can be interpreted as parallel to autonomous driving. Standardized in the ISO/TS 15066,
collaborative robot operations allow the machines (industrial manipulators, vehicles) installed in a shared workspace to move concurrently with human operators inside fenceless environments.  
Connected and collaborative robots are transforming industrial workspaces, such as assembly lines in the example of Fig. \ref{fed-2}(a), and represent a challenging ground for the development of decentralized FL tools. First, robots operate in increasingly complex and time-varying environments. Network design, paired with distributed learning tools, must consider the problem of continual learning (and periodic re-training) over URLLC communication links to track variations of data dynamics caused by changes of the workflow process.
Second, robots might perform distinct tasks, although these can be
considered as strongly related, targeting a common manufacturing process (assembly or dis-assembly tasks) and efficient workspace sharing. In such cases, opportunistic (ego) and collaborative approaches should be carefully assessed. For example, FL can boost the model training for some tasks, such as vision functions, in common to many robots.

To shed light on some of the above challenges, we resort to a typical mission critical control problem in Industry 4.0 workspaces. The goal is to plan the motions of a robotic manipulator based on the information about the positions of human operators sharing the workspace. Operator locations are classified using a ML model, described in Fig. \ref{fed-2}(a), that computes the human-robot distance $d$ and the direction of arrival (DOA) $\theta$ in real-time, using $6$ regions of interest as target classes. During the on-line workflow, the ML model can be trained/updated continuously by decentralized FL and consensus (cooperative learning loop). The sensing-decision-action loop is supervised by an industry standard programmable logic controller (PLC) that makes decisions about robot emergency stop, or trajectory replanning \cite{key-9}, based on the information about subject positions. Decentralized FL uses D2D links, runs in parallel to the sensing-decision-action loop, and thus takes some load off of the PLC network, whose resources must be reserved for robot motion control. All the robots support vision functions implemented by $3$ radars working in the $77-81$GHz band with a field-of-view of $120$deg each. Radars produce the raw data \cite{key-9} that are fed into the ML model.

\section{Results and discussions}

The example of Fig. \ref{fed-2}(b) analyses the performance of decentralized FL approaches compared with ego, i.e., opportunistic, learning, considering the application case previously described. In particular, the robots might choose to combine in groups of $15$ agents and implement FL over a D2D network, or rather act opportunistically, thus learning from local data only, disabling the D2D radio interface. The examples also highlight the impact of federated data distribution on convergence time and validation loss. When federated data is IID partitioned, each device uses $8\%$ of the full training dataset of 900 samples. In the more challenging non-IID scenario, only the $3\%$ of the data is used: batches contain examples for only $5$ of the $6$ output target classes, chosen randomly. 

D2D communications are organized into consecutive frames and use time-division multiple access (TDMA) scheduling. Frames have payload $1$ kB and TTI of $3$ ms, in line with URLLC \cite{bennis}. Notice that a frame drop (BLER is $10^{-9}$) causes the loss of a layer update, and an increase of the convergence time. FL and networking have been simulated on a virtual environment but using real data from the plant. The environment allows to deploy networked agents acting as virtual robots and learning over a configurable federated dataset (IID and non-IID). 

The simulations of Fig. \ref{fed-2}(b) quantify the average validation loss \cite{bennis} experienced by the deployed agents versus time. This allows to assess the time required by the consensus rounds, as relevant for real-time implementation. Gossip and diffusion FL approaches, described previously, are evaluated over a k-regular D2D network consisting of robots with $2$ neighbors each. Implementation of gossip is based on \cite{GOSSIPGRAD}: on each round, the agents randomly choose a single neighbor from the neighborhood, exchange the local ML models and adapt them by averaging, followed by SGD on local data. Diffusion FL implements the gradient negotiations described in Fig. \ref{exchange}: on each round the agents now exchange the local ML models \textit{and} the gradients $\nabla\mathbf{W}$ with a single neighbor, chosen again randomly. For both cases, the time span of an individual round consists of the transmission of the ML parameters in the assigned frames, the model adaptation and the SGD steps ($30$ms per round). Before D2D transmission, the ML model parameters are sparsified \cite{digvanalog} to limit the communication overhead, measured in kB per round.

Diffusion FL boosts the model training for robots possessing few data ($3\%$) and insufficient examples (non-IID). Compared with ego learning, cooperation gives smaller loss after a training period of $39$ s, that is enough to track workflow changes. Diffusion needs considerable communication overhead ($220$ kB per round) due to gradients exchange. Convergence time thus increases with the network size. Notice that small but sudden increases of the loss are observed on some rounds when averaging the neighbor models. These effects are more visible when the local models are trained with non-IID data and can be mitigated by learning rate optimization \cite{cfa}. Gossip requires twice lower number of frames ($92$ kB per round) but it is less effective over non-IID data. However, it is still useful for refining models trained with IID data as it improves ego approaches after $23$ s. Opportunistic learning converges fast as it does not utilize D2D communications, but experiences a large loss: it is a viable solution only for agents possessing enough data and performing specialized tasks that do not require much re-training.

\section{Conclusions and future directions}

In this article we explored opportunities and applications of FL in
networked and automated industrial systems, underpinned by D2D wireless
communications. Open problems and challenges have been discussed,
focusing on manufacturing and automotive B5G verticals. Decentralized FL enables the cooperative learning of ML models. It can be seamlessly integrated into the application
dependent sensing-decision-action loop within each automated entity
to improve knowledge discovery operations. Learning and re-training
continuously to follow the changing dynamics of the environment have
a profound impact on the networking co-design, capitalizing on model
sparsity and superposition properties of the wireless links.

Future research is expected to target increasingly complex mobile
environments characterized by heterogeneous devices cooperating to
learn distinct, but related, functions. The choice between opportunistic and cooperative behaviors will largely depend on the URLLC design. Emerging FL
tools are promising in cooperative automated driving, leveraging V2X
interactions, and in collaborative robotics for distributed learning
in complex and dynamic workflows. 

\section*{Appendix: Federated data and ML model}
	
As shown in Fig. \ref{fed-2}, the robots are equipped with $3$ Multiple-Input-Multiple-Output (MIMO) Frequency Modulated Continuous Wave (FMCW) radars, working in the $77$ GHz band. Radars implement a Time-Division (TD) MIMO system with $2$ transmit and $4$ receive antennas each, and a field-of-view of $120$ deg. During the on-line workflow, the distance $d$ and DOA $\theta$ information are classified by the agents using a trained ML model. The ML model is here trained to classify $6$ potential HR collaborative situations characterized by different HR distances and DOA ranges, corresponding to safe or unsafe conditions. 

Based on the above setup, a simplified database for testing, is available in the repository:\href{link}{http://dx.doi.org/10.21227/0wmc-hq36}. The database contains 4 main data structures, detailed as follows:

i) mmwave\_data\_test has dimension 900 x 256 x 63. Contains 900 FFT range-azimuth measurements of size 256 x 63: 256-point range samples corresponding to a max range of 11m (min range of 0.5m) and 63 angle bins, corresponding to AOA ranging from -75 to +75 degree. Data are used for testing (validation database). The corresponding labels are in label\_test. Each label (from 0 to 5) corresponds to one of the 6 positions (from 1 to 6) of the operator as detailed in the Fig. \ref{fed-2}.b.

ii) mmwave\_data\_train has dimension 900 x 256 x 63. Contains 900 FFT range-azimuth measurements used for training. The corresponding labels are in label\_train.

iii) label\_test with dimension 900 x 1, contains the true labels for test data (mmwave\_data\_test), namely classes (true labels) correspond to integers from 0 to 5. 

iv) label\_train with dimension 900 x 1, contains the true labels for train data (mmwave\_data\_train), namely classes (true labels) correspond to integers from 0 to 5. 

The implemented ML model takes as input the raw range-azimuth data (after background subtraction) of size $256\times63$ from the radars. As shown in Fig. \ref{fed-2}.a, it consists of $2$ trainable neural network layers of $25.000$ parameters. Decentralized FL uses gossip or diffusion methods (described previously) with SGD step size $\mu_{t}=0.025$.

%\vspace{-1.2cm}

\begin{IEEEbiographynophoto}{Stefano Savazzi}
is Researcher at the Consiglio Nazionale delle Ricerche (CNR) with
the Institute of Electronics, Computer and Telecommunication Engineering
(IEIIT). He was visiting researcher with the Uppsala University (2005) and the UCSD, San Diego (2009). He co-authored over 100 scientific publications. Research interests include signal processing, learning and networking design aspects for the Internet of Things and beyond 5G. He is the recipient of the Dimitris N. Chorafas Foundation Award. 
\end{IEEEbiographynophoto}

%\vspace{-1.2cm}
\balance 
\begin{IEEEbiographynophoto}{Monica Nicoli}
is an Associate Professor at Politecnico di Milano, with the Department of Management, Economics and Industrial Engineering. She was visiting researcher with ENI-Agip (Italy) in 1998-1999 and with Uppsala University (Sweden) in 2001. She co-authored over 100 scientific publications. Her research interests are in the area of statistical signal processing, with focus on wireless communications, localization and smart mobility. She is a recipient of the 1999 Marisa Bellisario Award, and a co-recipient of the 2014 IET Intelligent Transport Systems and the 2018 IEEE Statistical Signal Processing Workshop Best Paper Awards. 
\end{IEEEbiographynophoto}

\begin{IEEEbiographynophoto}{Mehdi Bennis}
is an Associate Professor at the Centre for Wireless Communications,
University of Oulu, Finland, an Academy of Finland Research Fellow
and head of the intelligent connectivity and networks/systems group
(ICON). His main research interests are in radio resource management,
heterogeneous networks, game theory and machine learning in 5G networks
and beyond. He was the recipient of several prestigious awards
including the 2015 Fred W. Ellersick Prize from the IEEE Communications
Society, the 2016 Best Tutorial Prize from the IEEE Communications
Society, the 2017 EURASIP Best paper Award for the Journal of Wireless
Communications and Networks, the all-University of Oulu award for
research and the 2019 IEEE ComSoc Radio Communications Committee Early
Achievement Award. 
\end{IEEEbiographynophoto}

%\vspace{-1.7cm}

\begin{IEEEbiographynophoto}{Sanaz Kianoush}
is a Postdoctoral researcher at the IEIIT institute of the CNR since November 2014. Visiting researcher at Aalto University in 2018, lecturer at Azad University Sary (Iran) in 2008.
Her research interests include statistical signal processing, machine
learning in communication systems, device-free radio localization. 
\end{IEEEbiographynophoto}

%\vspace{-1.7cm}

\begin{IEEEbiographynophoto}{Luca Barbieri}
is currently a Ph.D. candidate in Information Technology at DEIB department, Politecnico di Milano. His current research interests focus on machine learning and localization techniques for vehicular
and industrial networks. 
\end{IEEEbiographynophoto}

\end{document}